\documentclass[a4paper]{article}

\usepackage{INTERSPEECH2018}

\usepackage{hyperref}
\usepackage{url}
\usepackage{booktabs}       % professional-quality tables
\usepackage{graphicx}
\usepackage{color}

\newcommand{\xhdr}[1]{{\noindent\bfseries #1}.}

\newcommand{\cut}[1]{}

\title{Neural Transfer Learning for Cry-based Diagnosis of Perinatal Asphyxia }
\name{Charles C. Onu$^{1,2}$, Jonathan Lebensold$^{1,2}$, William L. Hamilton$^{1,3}$, Doina Precup$^{1,4}$}
\address{
   $^1$Mila - Qu\'ebec Artificial Intelligence Institute, McGill University \\
   $^2$ Ubenwa Health \\
   $^3$Facebook AI Research \\
  $^4$Google DeepMind
} 
\email{\{charles.onu@mail, jonathan.maloney-lebensold@mail, wlh@cs, dprecup@cs\}.mcgill.ca}

\begin{document}

\maketitle

\begin{abstract}
Despite continuing medical advances, the rate of newborn morbidity and mortality globally remains high, with over 6 million casualties every year. The prediction of pathologies affecting newborns based on their cry is thus of significant clinical interest, as it would facilitate the development of accessible, low-cost diagnostic tools\cut{ based on wearables and smartphones}. However, the inadequacy of clinically annotated datasets of infant cries limits progress on this task. This study explores a neural transfer learning approach to developing accurate and robust models for identifying infants that have suffered from perinatal asphyxia. In particular, we explore the hypothesis that representations learned from adult speech could inform and improve performance of models developed on infant speech. Our experiments show that models based on such representation transfer are resilient to different types and degrees of noise, as well as to signal loss in time and frequency domains.
\end{abstract}

\section{Introduction}

 Perinatal asphyxia---i.e., the inability of a newborn to breath spontaneously after birth---is responsible for one-third of newborn mortalities and disabilities worldwide \cite{who2017}. The high cost and expertise required to use standard medical devices for blood gas analysis makes it extremely challenging to conduct early diagnosis in many parts of the world.
In this work, we develop and analyze neural transfer models \cite{bengio2012deep} for predicting perinatal asphyxia based on the infant cry. We ask the question of whether such models could be more accurate and robust than previous approaches that primarily focus on classical machine learning algorithms due to limited data.

%Clinical researchers have shown that in babies experiencing certain pathologies like asphyxia, deafness, and autism, the cry patterns are significantly altered [cite, cite, cite]. Since traditional clinical tools require high cost and expertise, the use of infant cry for diagnosis of such conditions has long been an area of interest [cite, cite, cite] as it could lead to more cost-effective approaches - via smartphones, wearables or other handheld monitors - in resource-constrained areas.

Clinical research has shown that there exists a significant alteration in the crying patterns of newborns affected by asphyxia \cite{michelsson1977pain}. The unavailability of reasonably-sized clinically-annotated datasets limits progress in developing effective approaches for predicting asphyxia from cry. The Baby Chillanto Infant Cry database \cite{reyes2004system}, based on 69 infants, remains the only known available database for this task. Previous work using this data has mainly focused on classical machine learning methods or very limited capacity feed-forward neural networks \cite{reyes2004system,onu2014harnessing}.

We take advantage of freely available large datasets of adult speech to investigate a transfer learning approach to this problem using deep neural networks. 
In numerous domains (e.g., speech, vision, and text) transfer learning has led to substantial performance improvements by pre-training deep neural networks on some different but related task \cite{howard2018universal,oquab2014learning,karpathy2014large}. 
In our setting, we seek to transfer models trained on adult speech to improve performance on the relatively small Baby Chillanto Infant Cry dataset. 
Unlike newborns---whose cry is a direct response to stimuli---adults have voluntary control of their vocal organs and their speech patterns have been influenced, over time, by the environment. We nevertheless explore the hypothesis that there exists some underlying similarity in the mechanism of the vocal tract between adults and infants, and that model parameters learned from adult speech could serve as better initialization (than random) for training models on infant speech.

Of course, the choice of source task matters. The task on which the model is pre-trained should capture variations that are relevant to those in the target task.
For instance, a model pre-trained on a speaker identification task would likely learn embeddings that identify individuals, whereas a word recognition model would likely discover an embedding space that characterizes the content of utterances.
What kind of embedding space would transfer well to diagnosing perinatal asphyxia is not clear a priori.
For this reason, we evaluate and compare 3 different (source) tasks on adult speech: speaker identification, gender classification and word recognition.\cut{We also consider tasks that are not necessarily based on human speech, such as environmental sound classification.} We study how different source tasks affect the performance, robustness and nature of the learned representations for detecting perinatal asphyxia.

\xhdr{Key results} On the target task of predicting perinatal asphyxia, we find that a classical approach using support vector machines (SVM) represents a hard-to-beat baseline. Of the 3 neural transfer models, one (the word recognition task) surpassed the SVM's performance, achieving the highest unweighted average recall (UAR) of 86.5\%. By observing the response of each model to different degrees and types of noise, and signal loss in time- and frequency-domain, we find that all neural models show better {\em robustness} than the SVM.

\begin{figure*}[t]
  \centering
  \includegraphics[width=.8\textwidth]{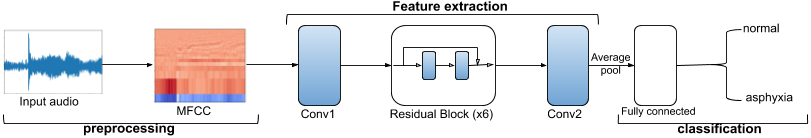}
  %\fbox{\rule[-.5cm]{0cm}{4cm} \rule[-.5cm]{4cm}{0cm}}
  \caption{Structure of learning pipeline. Weights of feature extraction stage were pre-loaded during transfer learning.}
\label{fig:model_diagram}
\end{figure*}

\section{Related Work}
\label{related_work}

\xhdr{Detecting pathologies from infant cry}
The physiological interconnectedness of crying and respiration has been long appreciated. Crying presupposes functioning of the respiratory muscles \cite{lagasse2005assessment}. In addition, cry generation and respiration are both coordinated by the same regions of the brain \cite{lester1990colic, zeskind2001analysis}. The study of how pathologies affect infant crying dates back to the 1970s and 1980s with the work of Michelsson et al \cite{michelsson1977pain, michelsson1977sound, michelsson2002cry}. Using spectrographic analysis, it was found that the cries of asphyxiated newborns showed shorter duration, lower amplitude, increased higher fundamental frequency, and significant increase in ``rising'' melody type.

\xhdr{The Chillanto Infant Cry database}
In 2004, Reyes et al.\@ \cite{reyes2004system} collected the Chillanto Infant Cry database with the objective of applying statistical learning techniques in classifying deafness, asphyxia, pain and other conditions. \cut{Cry recordings from 69 subjects were broken into 1 second segments to form a database of 1,389 examples. }The authors experimented with audio representations as linear predictive coefficients (LPC) and mel-frequency cepstral coefficients (MFCC), training a time delay neural network as the classifier. They achieved a precision and recall of 72.7\% and 68\%. Building on this work, Onu et al.\@ \cite{onu2014harnessing} improved the precision and recall to 73.4\% and 85.3\%, respectively, using support vector machines (SVM). It is worth noting that both works represent an overestimate of performance as authors split train/test set by examples, not by subjects.

\xhdr{Weight initialization and neural transfer learning}
 Modern neural networks often contain millions of parameters, leading to highly non-linear decision surfaces with many local optima. The careful initialization of the weights of these parameters has been a subject of continuous research, with the goal of increasing the probability of reaching a favorable optimum \cite{glorot2010understanding, he2015delving}. 
 %Some of these methods also provide guarantees on the properties of the weights throughout training (e.g. zero mean and unit variance). 
 Initialization-based transfer learning is based on the idea that instead of hand-designing a choice of random initialization, the weights from a neural network trained on similar data or task could offer better initialization. This pre-training could be done in an unsupervised \cite{erhan2010does} or supervised \cite{yosinski2014transferable, bengio2007greedy} manner. 
 %We implement the latter.
 
\begin{table}[b]
  \caption{Source tasks and corresponding datasets used in pre-training neural network. Size: number of audio files.}
  \label{table:source_task}
  \centering {\small
  \begin{tabular}{p{1.5cm}p{4.5cm}p{0.7cm}} %{lll}
    \toprule
    % \multicolumn{2}{c}{Part}                   \\
    % \cmidrule(r){1-2}
    Dataset   & Description & Size \\
    \midrule
    VCTK  & Speaker Identification. 109 English speakers reading sentences from newspapers.  & 44K \\
    \midrule
    SITW    & Gender classification. Speech samples from media of 299 speakers. & 2K \\
    \midrule
    Speech commands    &  Word recognition. Utterances from 1,881 speakers of a set of 30 words.  & 65K \\
    % Audio event recognition & AudioSet & Over weakly labelled clips of 527 audio classes. &  2M\\
    % Environmental sound classification & ESC-50 & Environment audio recordings organized into 50 semantic classes & 2K \\
    \bottomrule
  \end{tabular}}
\end{table}

\section{Methods}
In this section, we describe our approach to designing and evaluating transfer learning models for the detection of perinatal asphyxia in infant cry. We present the source tasks selected along with representative datasets. We further describe pre-processing steps, choice of model architectures as well as analysis of trained models.

\subsection{Tasks}
%\subsubsection{Source tasks}%
\subsubsection{Source tasks} We choose 3 source tasks --- speaker identification, gender classification, word recognition --- \cut{and 3 non-speech source tasks (audio event recognition, environmental sound classification)} with corresponding audio datasets: VCTK  \cite{veaux2017}, Speakers in the Wild (SITW)  \cite{mitchell2016}, and Speech Commands  \cite{warden2018} \cut{AudioSet \cite{gemmeke2017}, and ESC-50 \cite{piczak2015}}. Table \ref{table:source_task} briefly describes the datasets used for each task. 
\subsubsection{Target task: Perinatal asphyxia detection} Our target task is the detection of perinatal asphyxia from newborn cry. We develop and evaluate our models using the Chillanto Infant Cry Database. The database contains 1,049 recordings of normal infants and 340 cry recordings of infants clinically confirmed to have perinatal asphyxia. Audio recordings were 1-second long audio and sampled at frequencies between 8kHz to 16kHz with 16-bit PCM encoding.

\subsection{Pre-processing}
All audio samples are pre-processed similarly, to allow for even comparison between source tasks and compatibility with target task. Raw audio recordings are downsampled to 8kHz and converted to mel-frequency cepstral coefficients (MFCC). To do this, spectrograms were computed for overlapping frame sizes of 30 ms with a 10 ms shift, and across 40 mel bands. For each frame, only frequency components between 20 and 4000 Hz are considered. The discrete cosine transform is then applied to the spectrogram output to compute the MFCCs. The resulting coefficients from each frame are stacked in time to form a spatial ($40 \times 101$), 2D representation of the input audio.

%Examples of the MFCCs are shown in appendix \ref{appendix:spectrogram_examplesd}.

%, and across 40 Mel bands.

%. The discrete cosine transform is then applied to the spectrogram output to compute the MFCCs. The resulting values in each frame is stacked in time to form a spatial, 2D representation of the MFCCs. 

\subsection{Model Architecture and Transfer Learning}
 We adopt a residual network (ResNet) \cite{he2016deep} architecture with average pooling, for training. Consider a convolutional layer that learns a mapping function $F(x)$ of the input, parameterized by some weights. A residual block adds a shortcut or skip connection such that the output of the layer is the sum of $F(x)$ and the input $x$, i.e., $y = F(x) + x$. This structure helps control overfitting by allowing the network to learn the identity mapping $y=x$ as necessary and facilitates the training of even deeper networks.

ResNets represent an effective architecture for speech, achieving several state-of-the-art results in recent years \cite{tang2018deep}. To assure even comparison across source tasks, and to facilitate transfer learning, we adopt a single network architecture: the {\em res8} as in Tang et al.\@ \cite{tang2018deep}. The model takes as input a 2D MFCC of an audio signal, transforms it through a collection of 6 residual blocks (flanked on either side by a convolutional layer), employs average pooling to extract a fixed dimension embedding, and computes a k-way softmax to predict the classes of interest. Fig \ref{fig:model_diagram} shows the overall structure of our system. Each convolutional layer consists of 45, $3 \times 3$ kernels.

%In the work of \cite{tang2018deep}, 'res8' achieves an accuracy of 94\% on the speech commands dataset, which is 1\% less than the best model 'res15', but uses only half the number of parameters and 30 times fewer multiplies.

%\subsection{Transfer Learning}

We train the {\em res8} on each source task to achieve performance comparable with the state of the art. The learned model weights (except those of the softmax layer) are used as initialization for training the network on the Chillanto dataset. During this post-training, the entire network is tuned.

\subsection{Baselines}
We implement and compare the performance of our transfer models with 2 baselines. One is a model based on a radial basis function Support Vector Machine (SVM), similar to \cite{onu2014harnessing}. The other is a {\em res8} model whose initial weights are drawn randomly from a uniform Glorot distribution \cite{glorot2010understanding} i.e., according to $U(-k, k)$ where $k = \frac{\sqrt{6}}{n_i + n_o}$, and $n_i$ and $n_o$ are number of units in the input and output layers, respectively. This initialization scheme scales the weights in such a way that they are not too small to diminish or too large to explode through the network's layers during training.

%We implement and compare 4 methods for predicting perinatal asphyxia from infant cry: a support vector machine baseline (svm), a \emph{res8} network trained from random initialisation (res8-no-transfer), a \emph{res8} network pre-trained on source task 1 (res8-human-transfer), a \emph{res8} network pre-trained on source task 2 (res8-nonhuman-transfer).

%In order to obtain a pre-trained model for transfer, we train \emph{res8} on the Google speech commands dataset to classify the 10 typical keywords, unknown and silence, using the MFCC "images" as input.

%We further create a new instance of the res8 model, res8-transfer, for our task of classifying the presence or not of asphyxia, replacing only output layer to be a binary classifier. To transfer the learned model for our target task, we initialise res8-transfer using the weights from all but the fully-connected output layer of the pre-trained model. We then train the model from this start point. 

\subsection{Analysis}
    %\subsubsection{Performance}%
    \subsubsection{Performance} We evaluate the performance of our models on the target task by tracking the following metrics: sensitivity (recall on asphyxia class), specificity (recall on normal class), and the unweighted average recall (UAR). We use the UAR on the validation set for choosing best hyperparameter settings. The UAR is a preferred choice over accuracy since the classes in the Chillanto dataset are imbalanced.
    
    %We also visualise receiver-operating characteristics (ROC) curves for our models. ROC curves help crystallise the trade-off between sensitivity and specificity which is critical in any diagnostic system.
    
    %\subsubsection{Robustness to Additive Noise}%
    \subsubsection{Robustness}  %\textcolor{red}{to Additive Noise}} 
    \xhdr{Noise} We analyze our models for robustness to 4 different noise situations: Gaussian noise $\mathcal{N}(0,0.1)$, sounds of children playing, dogs barking and sirens. In each case, we insert the noise in increasing magnitude to the test data and monitor the impact on classification performance of the model.
    
    \xhdr{Audio length} We also evaluate the response of each model to varying lengths of audio, since in the real-world a diagnostic system must be able to work with as much data as is available. To achieve this, we test the models on increasing lengths of the test data, starting from 0.1s to the full 1s segment, in 0.1 increments.
    
    \xhdr{Frequency response} The response of the models to variations in frequency domain is important as this could reveal underlying characteristics of the data. We know as well that perinatal asphyxia alters the frequency patterns in cry. To discover what range of frequencies are most sensitive in detecting perinatal asphyxia, we conduct an ablation exercise where features extracted from a different filterbanks in the MFCC are zeroed out. We measure the response of our models by monitoring the drop in performance for the frequency ranges in each mel-filterbank.

\begin{figure}
\includegraphics[width=0.42\textwidth]{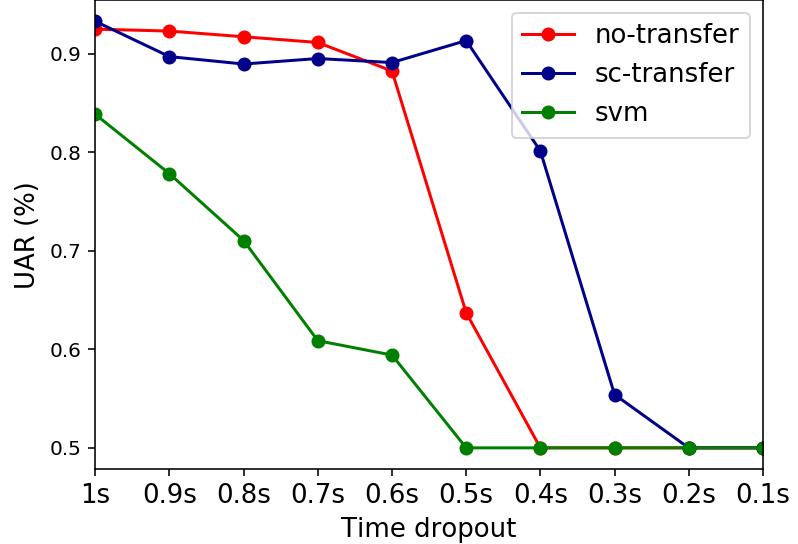}
  %\fbox{\rule[-.5cm]{0cm}{4cm} \rule[-.5cm]{4cm}{0cm}}
\caption{Audio length analysis highlighting the impact of using shorter amounts of input audio on UAR performance.}
\label{fig:time_analysis}
%   Left: noise distributed as Gaussian random variable of $\mathcal{N}[0,1]$. Right: sounds of children playing obtained from \cite{salamon2014}.}
\end{figure}

\begin{figure}
  \includegraphics[width=0.45\textwidth]{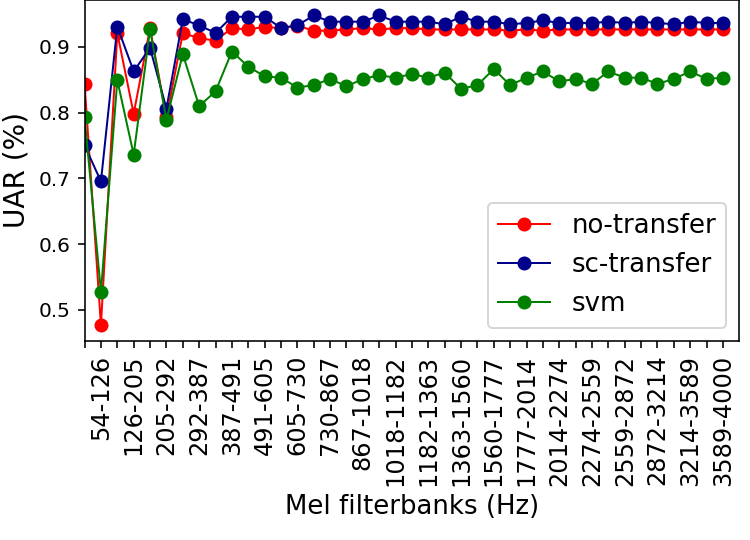}
  %\fbox{\rule[-.5cm]{0cm}{4cm} \rule[-.5cm]{4cm}{0cm}}
  \caption{Frequency response analysis of the relative importance of different Mel filterbanks on UAR performance. Each point represents the performance after removing the corresponding Mel filterbank.}
%   Left: noise distributed as Gaussian random variable of $\mathcal{N}[0,1]$. Right: sounds of children playing obtained from \cite{salamon2014}.}
\label{fig:freq_analysis}
\end{figure}
    %\subsubsection{MFCC Embeddings}%
    \subsubsection{MFCC Embeddings} In order to further investigate the nature of the embedding learned by each model, we apply principal component analysis (PCA) to the learned final-layer embeddings for all models \cite{jolliffe2011principal}.
    %PCA finds the axes of maximum linear variations (the principal components) in the given input data set using the method of singular value decomposition. 
    By applying PCA, we hope to gain insight on the extent to which the embedding space captures unique information.
    
    %In order to visualize the nature of the embedding learned by each model, we apply a combination of principal component analysis (PCA) \cite{pca} and locally linear embedding (LLE) \cite{lle} to compress the n-dimensional embedding to 2 dimensions. We first use singular value decomposition to get the top principal components which explain up to 95\% of the data. We then employ LLE, a non-linear dimensionality reduction technique to compress the PCs to 2 dimensions.

    % \subsubsection{Error analysis}
    % visualise probability distributions of different acoustic properties of misclassified samples.

\begin{table}[th]
  \caption{Performance -- mean (standard error) -  of different models in predicting perinatal asphyxia. \cut{%Will: I think this is clear enough and we need the space, but not certain.
  The models are Support Vector Machine (SVM), a res8 network trained from random initialization (res8-no-transfer), and res8 networks pretrained on speaker identification (speaker-id-transfer), gender identification (gender-id-transfer), word recognition (word-rec-transfer), audio event recognition (event-rec-transfer) and environmental sound classification (sound-id-transfer) tasks.}}
  \label{table:performance}
  \centering {\small 
  \begin{tabular}{lllll}
\toprule
   Model    &  UAR \%       &     Sensitivity \%            &   Specificity \%       \\
\midrule
SVM & 84.4 (0.4)  & 81.6 (0.7) & 87.2 (0.2) \\
% \midrule
% Transfer models & & & & \\
% \cmidrule(r){1-1}
no-transfer   &    80.0 (2.5) &            71.8 (5.8) &            88.1 (0.8) \\
sc-transfer   &    \textbf{86.5 (1.1)} &            \textbf{84.1 (2.2)} &            \textbf{88.9 (0.4)} \\
sitw-transfer &    81.1 (1.7) &            72.7 (3.5) &            89.5 (0.2) \\
vctk-transfer &    80.7 (1.0) &            72.2 (2.1) &            89.1 (0.3) \\
\bottomrule
\end{tabular}}
\end{table}

\begin{figure*}[t]
  \centering
  \includegraphics[width=0.8\textwidth]{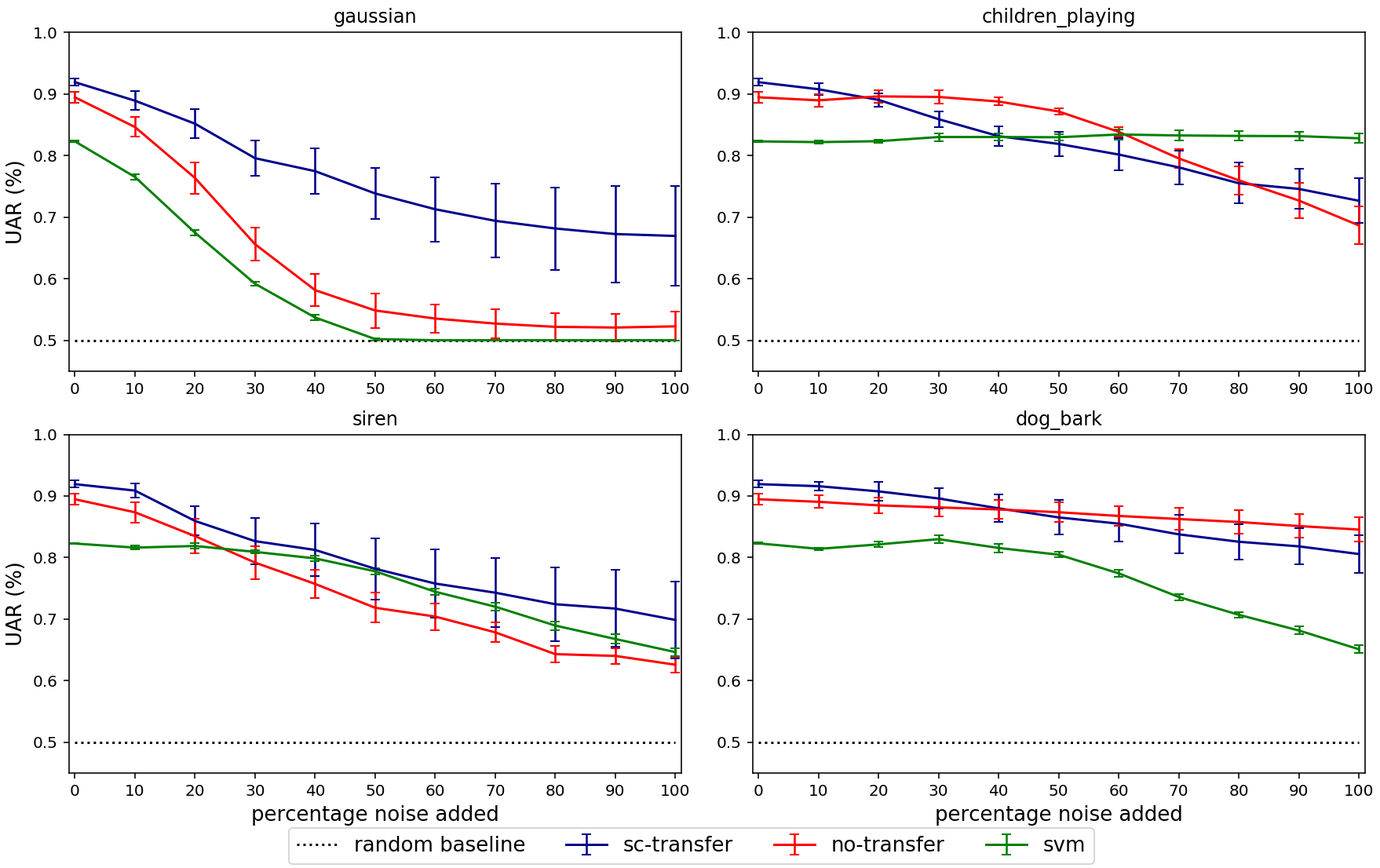}
  %\fbox{\rule[-.5cm]{0cm}{4cm} \rule[-.5cm]{4cm}{0cm}}
  \caption{Performance of models under different noise conditions. }
  \label{fig:noise_analysis}
%   Left: noise distributed as Gaussian random variable of $\mathcal{N}[0,1]$. Right: sounds of children playing obtained from \cite{salamon2014}.}
\end{figure*}

\section{Experiments}

\subsection{Training details}
There were a total of 1,389 infant cry samples (1,049 normal and 340 asphyxiated) in the Chillanto dataset. The samples were split into training, validation and test sets, with a 60:20:20 ratio, and under the constraint that samples from the same patients were placed in the same set.

Each source task was trained, fine-tuning hyperparameters as necessary to obtain performance comparable with the literature. For transfer learning on the target task, models were trained for 50 epochs using stochastic gradient descent with an initial learning rate of 0.001 (decreasing to 0.0001 after 15 epochs), a fixed momentum of 0.9, batch size of 50, and hinge loss function. We used a weighted balanced sampling procedure for mini-batches to account for class imbalance. We also applied data augmentation via random time-shifting of the audio recordings. Both led to up to 7\% better UAR scores when training source and target models.

\subsection{Performance on source tasks}
Our model architecture achieved accuracies of 94.8\% on word recognition task (Speech Commands), 91.9\% on speaker identification (VCTK) and 90.2\% on gender classification (SITW). These results are comparable to previous work. See \cite{tang2018deep,arik2018neural} for reference \footnote{SITW to our knowledge has not been used for gender classification, even though this data is available}.

\cut{
\begin{table}[th]
  \caption{Source task performance}
  \label{table:source_task_performance}
  \centering
  \begin{tabular}{p{2.cm}p{2.5cm}p{1.5cm}} %
    \toprule
    Dataset     & Random Baseline (\%) & Test Acc. (\%) \\
    \midrule
    VCTK    & 0.9 & 22.9 \\
    SITW     & 50 & 87.1 \\
    Speech Comm.    & 8.3 & 94.2 \\
    AudioSet    & 0.2 & 35.9 \\
    ESC-50     & 2 & 76.8 \\
    \bottomrule
  \end{tabular}
\end{table}
}

\subsection{Performance on target task}
Table \ref{table:performance} summarizes the performance of all models on the target task. The best performing model was pre-trained on the word recognition task (sc-transfer) and attained a UAR of 86.5\%. This model also achieves the highest sensitivity and specificity 84.1\% and 88.9\% respectively. All other transfer models performed better than {\em no-transfer }, suggesting that transfer learning resulted in better or at least as good an initialization. The SVM was the second best performing model and had the lowest variance among all models in its predictions.

%   \begin{tabular}{lllll}
%     \toprule
%     % \multicolumn{2}{c}{Part}                   \\
%     % \cmidrule(r){1-2}
%     Model     & Sensitivity (\%)     & Specificity (\%) & Precision (\%) & F1 Score (\%) \\
%     \midrule
%     SVM & 81.1  & 87.3  & 56.6 & 66.7  \\
%     res8-no-transfer     & 73.6 & 89.6 & 59.0 & 65.5   \\
%     res8-wr-transfer (speech)     & \textbf{92.5} & \textbf{91.5} & \textbf{69.0} & \textbf{79.0}  \\
%     \bottomrule
%   \end{tabular}

\subsection{Robustness Analysis}
In most cases, our results suggest that neural models have overall increased robustness. We focused on the top transfer model {\em sc-transfer}, {\em no-transfer} and the SVM. Figure \ref{fig:noise_analysis}, shows the response of the models to different types of noise, revealing that in all but one case the neural models degrade slower than the SVM.  Results from Figure \ref{fig:time_analysis} suggest that the neural models are also capable of high UAR scores for short audio lengths, with {\em sc-transfer} maintaining peak performance when evaluated on only half (0.5s) of the test signals.

From our analysis of the models' responses to filterbank frequencies (Figure \ref{fig:freq_analysis}), we observe that (i) the performance of all models (unsurprisingly) only drops in the range of the fundamental frequency of infant cries, i.e. up to 500Hz \cite{daga2011acoustical} and (ii) {\em sc-transfer} again is the most resilient model across the frequency spectrum.

\subsection{Visualization of embeddings} \label{appendix:pca_analysis}

Figure 5 shows cumulative variance explained by the principal components (PC) of the neural model embeddings. Whereas in {\em no-transfer}, the top 2 PCs explain nearly all variance in the data (91\%), in {\em sc-transfer} they represent only 52\%---suggesting that the neural transfer leads to an embedding that is intrinsically higher dimensional and richer than the {\em no-transfer} counterpart.

\begin{figure}[t]
\label{fig:pca_cum_var}
  \centering
  \includegraphics[width=.4\textwidth]{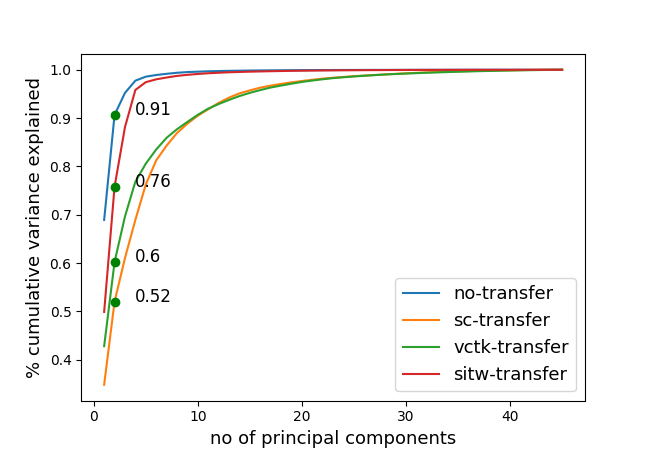}
  %\fbox{\rule[-.5cm]{0cm}{4cm} \rule[-.5cm]{4cm}{0cm}}
  \caption{Cumulative variance explained by all principal components (left) and the top 2 principal components on the Chillanto test data (right) based on embeddings of {\em no-transfer} model.}
\end{figure}

\section{Conclusion and Discussion}
We compared the performance of a residual neural network (ResNet) pre-trained on several speech tasks in classifying perinatal asphyxia. Among the transfer models, the one based on a word recognition task performed best, suggesting that the variations learned for this task are most analogous and useful to our target task. The support vector machine trained directly on MFCC features proved to be a strong benchmark, and if variance in predictions was of concern, a preferred model. The SVM, however, was clearly less robust to pertubations in time- and frequency-domains than the neural models. This work reinforces the modelling power of deep neural networks. More importantly, it demonstrates the value of a transfer learning approach to the task of predicting perinatal asphyxia from the infant cries---a task of critical relevance for improving the accessibility of pediatric diagnostic tools.

% TODO: need to loop the discussion back better to the problem and to adult speech transfer. 
% Someone who reads only the conclusion should be reminded of the problem (both health and computational), the solution and why it is a good solution.

\bibliographystyle{IEEEtran}
\bibliography{main}

\end{document}